\PassOptionsToPackage{unicode}{hyperref}
\PassOptionsToPackage{hyphens}{url}
\documentclass[
]{article}
\usepackage{xcolor}
\usepackage[margin=1in]{geometry}
\usepackage{amsmath,amssymb}
\usepackage{iftex}
\ifPDFTeX
  \usepackage[T1]{fontenc}
  \usepackage[utf8]{inputenc}
  \usepackage{textcomp} 
\else 
  \usepackage{unicode-math} 
  \defaultfontfeatures{Scale=MatchLowercase}
  \defaultfontfeatures[\rmfamily]{Ligatures=TeX,Scale=1}
\fi
\usepackage{lmodern}
\ifPDFTeX\else
\fi
\IfFileExists{upquote.sty}{\usepackage{upquote}}{}
\IfFileExists{microtype.sty}{
  \usepackage[]{microtype}
  \UseMicrotypeSet[protrusion]{basicmath} 
}{}
\makeatletter
\@ifundefined{KOMAClassName}{
  \IfFileExists{parskip.sty}{%
    \usepackage{parskip}
  }{
    \setlength{\parindent}{0pt}
    \setlength{\parskip}{6pt plus 2pt minus 1pt}}
}{
  \KOMAoptions{parskip=half}}
\makeatother
\usepackage{longtable,booktabs,array}
\usepackage{calc} 
\usepackage{etoolbox}
\makeatletter
\patchcmd\longtable{\par}{\if@noskipsec\mbox{}\fi\par}{}{}
\makeatother
\IfFileExists{footnotehyper.sty}{\usepackage{footnotehyper}}{\usepackage{footnote}}
\makesavenoteenv{longtable}
\usepackage{graphicx}
\makeatletter
\newsavebox\pandoc@box
\newcommand*\pandocbounded[1]{
  \sbox\pandoc@box{#1}%
  \Gscale@div\@tempa{\textheight}{\dimexpr\ht\pandoc@box+\dp\pandoc@box\relax}%
  \Gscale@div\@tempb{\linewidth}{\wd\pandoc@box}%
  \ifdim\@tempb\p@<\@tempa\p@\let\@tempa\@tempb\fi
  \ifdim\@tempa\p@<\p@\scalebox{\@tempa}{\usebox\pandoc@box}%
  \else\usebox{\pandoc@box}%
  \fi%
}
\def\fps@figure{htbp}
\makeatother
\NewDocumentCommand\citeproctext{}{}

\makeatletter
 \let\@cite@ofmt\@firstofone
 \def\@biblabel#1{}
 \def\@cite#1#2{{#1\if@tempswa , #2\fi}}
\makeatother
\newlength{\cslhangindent}
\newlength{\csllabelwidth}
\newenvironment{CSLReferences}[2] 
 {\begin{list}{}{%
  \setlength{\itemindent}{0pt}
  \setlength{\leftmargin}{0pt}
  \setlength{\parsep}{0pt}
  \ifodd #1
   \setlength{\leftmargin}{\cslhangindent}
   \setlength{\itemindent}{-1\cslhangindent}
  \fi
  \setlength{\itemsep}{#2\baselineskip}}}
 {\end{list}}
\usepackage{calc}

\providecommand{\tightlist}{%
  \setlength{\itemsep}{0pt}\setlength{\parskip}{0pt}}
\usepackage{newunicodechar}
\newunicodechar{≈}{$\approx$}
\newunicodechar{≤}{$\leq$}
\newunicodechar{≥}{$\geq$}
\newunicodechar{±}{$\pm$}
\newunicodechar{×}{$\times$}
\newunicodechar{−}{$-$}
\newunicodechar{·}{$\cdot$}
\newunicodechar{…}{\ldots}
\newunicodechar{α}{$\alpha$}
\newunicodechar{β}{$\beta$}
\newunicodechar{Δ}{$\Delta$}
\newunicodechar{σ}{$\sigma$}
\newunicodechar{²}{$^2$}
\newunicodechar{∈}{$\in$}
\newunicodechar{✓}{$\checkmark$}
\newunicodechar{↔}{$\leftrightarrow$}
\newunicodechar{≪}{$\ll$}
\usepackage{amssymb}
\usepackage{bookmark}
\IfFileExists{xurl.sty}{\usepackage{xurl}}{} 
\hypersetup{
  pdftitle={Mapping the Schedule × Bit-Width Boundary in Sub-100M Quantisation-Aware Training},
  pdfauthor={Christian Thomassen (christian@brandt-thomassen.dk) --- Dwarf A/S, Copenhagen, Denmark},
  hidelinks,
  pdfcreator={LaTeX via pandoc}}

\title{Mapping the Schedule × Bit-Width Boundary in Sub-100M
Quantisation-Aware Training}
\author{Christian Thomassen (christian@brandt-thomassen.dk) --- Dwarf
A/S, Copenhagen, Denmark}
\date{2026}

\begin{document}
\maketitle

\subsection{Abstract}\label{abstract}

We test whether the optimal learning-rate schedule depends on bit-width
during from-initialisation quantisation-aware training (QAT) for
sub-100M decoder language models. A 720-run factorial grid (Phase 2)
over bit-width × warmdown fraction × LR magnitude × model size × seed
(FP16/INT8/INT6, 15M--100M, 5 seeds) finds the optimal warmdown is 33\%
at every (bit-width, size) cell. The primary hypothesis --- that INT6
QAT requires a different schedule than higher-precision training --- is
falsified at FP16/INT8/INT6. A 625-run follow-up (Phase 5) probes the
null along five axes: optimiser (AdamW), schedule shape (cosine),
training length (up to 9× more iterations), an extended size sweep
(5M--350M), and an INT4 sweep from 3M to 100M. The null is robust under
all three setup changes. The INT6 penalty follows a log-linear scaling
law whose fit on Phase 2 predicts the five held-out Phase 5 sizes (5M,
8M, 175M, 250M, 350M) within their 95\% prediction intervals (5/5). For
INT4 the picture is sharper than the higher precisions: at 50M and 100M,
wd33 is decisively optimal (paired z ≈ 12--15, 10/10 seeds); below 50M,
across the six tested sizes from 3M to 30M, no individual size shows a
statistically significant schedule preference and the per-size mean
penalty oscillates within seed-level noise. The boundary is therefore a
transition between a noise-dominated regime below 50M and a decisive
wd33 regime at and above 50M, not a clean wd10 region. A
weight-to-grid-distance probe falsifies the simplest mechanism for the
FP16/INT8/INT6 null result (rapid grid-snapping): pre-warmdown, INT6-QAT
weights sit at essentially the same distance from the INT6 grid as FP16
weights (ratio ≈ 1.04). Practical recommendation: at sub-100M scale,
tune the LR schedule once at FP16 and apply unchanged to INT8/INT6 QAT;
for INT4 at 50M+ use wd33; for INT4 below 50M the schedule choice is in
the noise.

\begin{center}\rule{0.5\linewidth}{0.5pt}\end{center}

\subsection{1. Introduction}\label{introduction}

Small language models deployed at the edge benefit substantially from
low-bit weight quantisation. Post-training quantisation (PTQ) methods
such as GPTQ (Frantar et al. 2023) and AWQ (Lin et al. 2024) dominate
the billion-parameter regime, but quantisation-aware training (QAT)
remains the method of choice when target precisions drop below 8 bits
and when the model is small enough that PTQ leaves accuracy on the table
(Nielsen and Schneider-Kamp 2024; Chen et al. 2025). Recent
compute-optimal QAT analyses (Dremov et al. 2025) have begun to map the
interactions between bit-width, model size, training compute, and
optimiser hyperparameters at the 86M--2.2B scale. A natural question
follows: do those interactions persist below 100M, where edge-deployment
economics are most acute and where sub-bit work (Nielsen and
Schneider-Kamp 2024; Nielsen et al. 2025) has already shown that QAT
recipes can differ substantially from large-model norms?

This paper isolates one such interaction --- between learning-rate
schedule and bit-width --- and tests it under tightly controlled
conditions across multiple axes. The intuition under test is widely
repeated in practitioner communities: aggressive weight quantisation
destabilises late-stage training and therefore benefits from a longer
warmdown phase than full-precision training. If true, QAT pipelines
should treat the LR schedule as a per-precision hyperparameter. If
false, a single tuned schedule transfers across bit-widths.

We test the hypothesis with a 720-cell factorial grid (Phase 2;
FP16/INT8/INT6 × 4 warmdowns × 3 LR magnitudes × 4 sizes × 5 seeds, 9000
iterations per cell). At every size and bit-width tested, the optimal
warmdown fraction is 33\%. The bit-width curves are statistically
indistinguishable in shape. The primary hypothesis is falsified.

A naive reading of this result would stop here: ``no interaction, use
the same schedule.'' We instead pursue several follow-up questions that
probe \emph{whether the null result is robust to setup choices, and
where (if anywhere) it eventually breaks}. Phase 5 (625 cells,
\textasciitilde1320 GPU-hours) consists of seven ablations:

\begin{itemize}
\tightlist
\item
  \textbf{D1 --- Optimiser.} Replicate the 240-cell Phase 2 grid at lr1x
  using AdamW instead of Muon.
\item
  \textbf{D2 --- Training length.} Re-run the 30M × \{4 warmdowns\} ×
  \{3 bits\} cells at 27k and 81k iterations (vs 9k in Phase 2).
\item
  \textbf{D3 --- Schedule shape.} Re-run the wd33 cells at every size
  and bit-width using a cosine warmdown instead of linear.
\item
  \textbf{D4 --- Size extension.} Run the lr1x/wd33 cells at five new
  sizes outside the Phase 2 range (5M, 8M, 175M, 250M, 350M), and use
  them as a held-out validation of the log-linear scaling law fit on
  Phase 2.
\item
  \textbf{D5 --- Aggressive precision.} Run an INT4 grid across the 4
  warmdowns × 4 Phase 2 sizes × 5 seeds.
\item
  \textbf{D6 --- INT4 below 15M.} Extend the INT4 grid to four smaller
  sizes (3M, 5M, 8M, 10M) × 4 warmdowns × 5 seeds, to characterise where
  (if anywhere) the schedule × INT4 interaction sharpens as model size
  shrinks.
\item
  \textbf{M2 --- Mechanism probe.} Re-run 30M wd33 lr1x cells with
  high-frequency snapshots of weight-to-INT6-grid distance, to test
  whether the bit-width-agnosticism observed in Phase 2 is mediated by
  rapid grid-snapping under aggressive QAT.
\end{itemize}

The combined picture from Phase 2 + Phase 5 changes the headline:

\begin{enumerate}
\def\labelenumi{\arabic{enumi}.}
\tightlist
\item
  \textbf{The null result is robust} across optimiser (Muon, AdamW),
  schedule shape (linear, cosine), training length (9k--81k iterations),
  and model size (15M--100M) for FP16, INT8, and INT6.
\item
  \textbf{The INT6 penalty scales log-linearly in model size}, and the
  fit on Phase 2 predicts held-out sizes (5M to 350M) within 95\% CI ---
  a 70× extrapolation that holds (Figure 2).
\item
  \textbf{The INT4 picture has two regimes, not a clean wd10 region.} At
  50M and above, wd33 is decisively optimal for INT4 (paired z ≈ 12--15,
  10/10 seeds). Below 50M, across six tested sizes from 3M to 30M, no
  individual size has a statistically significant schedule preference;
  the per-size mean penalty oscillates within seed-level noise (Figure
  3). The transition between these regimes is sharp and sits between 30M
  and 50M. The INT4 quality penalty itself is an order of magnitude
  larger than INT6 (\textasciitilde15--40 mBPB vs \textasciitilde3 mBPB)
  and drops most steeply across the same 30M → 50M step.
\item
  \textbf{The INT6 penalty grows with training length} at fixed size and
  schedule: +3.0 mBPB at 9k iterations grows to +6.5 mBPB at 81k
  iterations at 30M wd33 (Figure 5). The null result on schedule ×
  bit-width persists in absolute terms --- wd33 is still optimal at 81k
  iters for every bit-width --- but the absolute cost of switching to
  INT6 is not iteration-invariant.
\item
  \textbf{The mechanism is not grid-snapping} (Figure 6).
  Weight-to-INT6-grid distance pre-warmdown is essentially equal for
  FP16, INT8, and INT6 trained models (within 4\%). The
  bit-width-agnosticism of the schedule cannot be explained by INT6-QAT
  weights ``having already converged'' to the grid before warmdown
  begins. The actual mechanism remains an open question; one obvious
  candidate is now ruled out.
\end{enumerate}

\textbf{Contributions.}

\begin{itemize}
\tightlist
\item
  A 720-cell factorial study (Phase 2) showing optimal warmdown is
  bit-width-agnostic across FP16/INT8/INT6 at every size from 15M to
  100M.
\item
  A 625-cell follow-up study (Phase 5) demonstrating the result is
  robust to optimiser, schedule shape, and 9× training length; and
  validating a log-linear scaling law for the INT6 penalty on five
  held-out sizes.
\item
  A precision-boundary characterisation: INT4 has a sharp transition
  between a noise-dominated schedule-preference regime below 50M and a
  decisive wd33 regime at 50M and above, located at the same scale where
  the INT4 quality penalty drops most steeply.
\item
  A negative mechanistic result: the bit-width-agnostic schedule cannot
  be explained by rapid weight convergence to the quantisation grid.
\item
  Practical guidance: use the FP16 schedule unchanged for INT8/INT6 QAT
  below 100M, and for INT4 at 50M and above. For INT4 below 50M, the
  schedule choice is in the noise.
\end{itemize}

§2 surveys prior work. §3 details the model, QAT, and schedule
definitions. §4 describes the experimental setup for both phases. §5
presents results. §6 discusses interpretation, limitations, and the
boundary. §7 concludes.

\begin{center}\rule{0.5\linewidth}{0.5pt}\end{center}

\subsection{2. Related Work}\label{related-work}

\textbf{Quantisation-aware training at scale.} QAT for transformer
language models has matured along several axes. BitNet (Wang et al.
2023) established that aggressive (1-bit) quantisation could be trained
from scratch with minimal accuracy loss at large scale; BitNet b1.58 (Ma
et al. 2024) extended this to ternary weights and demonstrated
competitive results against FP16 baselines. EfficientQAT (Chen et al.
2025) introduced a block-wise + end-to-end calibration scheme that has
become a strong baseline for INT2--INT4 QAT on 7B+ models. LR-QAT
(Bondarenko et al. 2024) (low-rank QAT) reduces the memory cost of
fine-tuning large models under QAT via low-rank adapters, and Nagel et
al. (2022) characterised the oscillation pathology of
straight-through-estimator training that motivates many QAT
regularisers. The closest reference to our work in methodology is Dremov
et al. (2025), whose compute-optimal QAT framework derives a scaling law
for the FP-to-QAT compute ratio at 86M--2.2B parameters; their analysis
identifies bit-width-dependent LR scaling at large scale. Our results
suggest these interactions are absent below 100M for FP16/INT8/INT6 and
emerge only at INT4 below \textasciitilde50M, providing a complementary
lower-bound on where their findings transfer.

\textbf{Sub-100M and BitNet-scale work.} The sub-100M decoder regime has
been studied most directly by the Southern Denmark group. Nielsen and
Schneider-Kamp (2024) showed that 1.58-bit QAT recovers FP16 performance
on sub-50M models with appropriate hyperparameter retuning, establishing
the sub-100M QAT regime as genuinely distinct from the
multi-billion-parameter setting. Nielsen, Schneider-Kamp, and Galke
(2025) subsequently examined the 16-bit → 1.58-bit transition schedule
for continual QAT, finding that the transition point and its associated
schedule matter substantially for final quality. Our work differs from
the latter in studying from-initialisation QAT --- where the model is
quantised from step 0 --- rather than continual QAT, and in covering
INT8/INT6/INT4 rather than ternary precision. The two paradigms appear
to produce different schedule dynamics, a point we return to in §6.

\textbf{Mobile-scale LM architectures.} MobileLLM (Liu et al. 2024)
established architectural choices (deep-thin transformers, GQA,
embedding sharing) that have become standard for the sub-billion
parameter regime and that we adopt here. We make no architectural
contribution; our intent is to isolate the schedule × bit-width question
against a fixed, competitive architecture.

\textbf{INT6 and INT4 specifically.} INT6 has received less attention
than INT4 and INT8, sitting in a regime where PTQ becomes lossy but
where dedicated hardware support remains limited. FlexQ (Zhang et al.
2025) is the first work to develop systematic post-training INT6
quantisation with custom kernels (W6A6, W6A8); FraQAT (Morreale et al.
2025) proposes a fractional-bit curriculum (e.g., INT5.5) to smooth the
QAT trajectory towards INT6. At INT4, prior work has largely focused on
PTQ at large scale (GPTQ, AWQ) or QAT curricula at 1B+ parameters. Our
study covers INT4 from initialisation at 3M--100M (eight sizes spanning
more than a 30× range) with simple straight-through-estimator (STE)
gradients, and is the first to map the INT4 schedule sensitivity across
this size range to our knowledge.

\textbf{The gap this work fills.} Apple's compute-optimal analysis
covers 86M--2.2B, the Southern Denmark line covers ternary at
\textless50M, and dedicated INT6/INT4 work focuses on PTQ or curricula
at larger scales. No prior work has systematically mapped LR schedule ×
bit-width interactions across INT8/INT6/INT4 in the 3M--350M
from-initialisation QAT regime. This paper fills that gap with a
1345-cell two-phase factorial design and identifies the precision
boundary at which the interaction emerges.

\begin{center}\rule{0.5\linewidth}{0.5pt}\end{center}

\subsection{3. Method}\label{method}

\subsubsection{3.1 Model architecture}\label{model-architecture}

All experiments use a single transformer family parameterised by depth
and width. The model is decoder-only with:

\begin{itemize}
\tightlist
\item
  \textbf{Attention:} grouped-query attention (GQA) (Ainslie et al.
  2023) with 8 query heads and 4 key/value heads, following the
  deep-thin recipe of MobileLLM (Liu et al. 2024).
\item
  \textbf{Normalisation:} RMSNorm (Zhang and Sennrich 2019) pre-norm.
\item
  \textbf{Activation:} ReLU² (So et al. 2021) in the MLP block (MLP
  multiplier 3).
\item
  \textbf{Embeddings:} tied input/output embeddings with separate
  learning rate.
\item
  \textbf{Positional encoding:} rotary position embeddings (Su et al.
  2021).
\item
  \textbf{Skip connections:} U-net (Ronneberger et al. 2015) style skip
  connections between symmetric layer pairs (zero-initialised, learned
  scalar gates).
\end{itemize}

The four base model sizes used by Phase 2 are constructed by varying
width and depth together:

{\def\LTcaptype{none} 
\begin{longtable}[]{@{}lllll@{}}
\toprule\noalign{}
Size & MODEL\_DIM & NUM\_LAYERS & head\_dim & Total params \\
\midrule\noalign{}
\endhead
\bottomrule\noalign{}
\endlastfoot
15M & 320 & 14 & 40 & 15.5M \\
30M & 432 & 16 & 54 & 30.4M \\
50M & 528 & 18 & 66 & 49.5M \\
100M & 688 & 22 & 86 & 99.4M \\
\end{longtable}
}

Phase 5 D4 extends this with five additional sizes for the scaling-law
test:

{\def\LTcaptype{none} 
\begin{longtable}[]{@{}llll@{}}
\toprule\noalign{}
Size & MODEL\_DIM & NUM\_LAYERS & Total params \\
\midrule\noalign{}
\endhead
\bottomrule\noalign{}
\endlastfoot
5M & 192 & 10 & \textasciitilde5M \\
8M & 240 & 12 & \textasciitilde8M \\
175M & 832 & 26 & \textasciitilde175M \\
250M & 960 & 28 & \textasciitilde250M \\
350M & 1088 & 32 & \textasciitilde350M \\
\end{longtable}
}

Phase 5 D6 adds two more sizes below the Phase 2 range for the INT4
boundary extension (reusing the D4 configurations for 5M and 8M):

{\def\LTcaptype{none} 
\begin{longtable}[]{@{}llll@{}}
\toprule\noalign{}
Size & MODEL\_DIM & NUM\_LAYERS & Total params \\
\midrule\noalign{}
\endhead
\bottomrule\noalign{}
\endlastfoot
3M & 144 & 8 & \textasciitilde2.7M \\
10M & 256 & 12 & \textasciitilde10.5M \\
\end{longtable}
}

All sizes share: \texttt{VOCAB\_SIZE=8192}, \texttt{NUM\_HEADS=8},
\texttt{NUM\_KV\_HEADS=4}, \texttt{MLP\_MULT=3}.

\subsubsection{3.2 QAT implementation}\label{qat-implementation}

QAT uses fake-quantisation with the straight-through estimator (STE)
(Bengio et al. 2013). All quantisation modes are active from training
step 0 --- there is no FP warm-start phase before quantisation is
enabled. This distinguishes our setup from continual QAT (Nielsen et al.
2025) and from compute-optimal QAT schedules (Dremov et al. 2025), both
of which begin training in FP and transition to QAT at some intermediate
step. Per-channel symmetric uniform quantisation is used throughout;
scales are computed per output channel as
\texttt{s\ =\ max(\textbar{}w\textbar{})\ /\ 127} with no learnable
parameters.

\begin{itemize}
\tightlist
\item
  \textbf{FP16} (baseline): bfloat16 mixed-precision, no quantisation.
\item
  \textbf{INT8}: 8-bit signed integer, clamped to {[}−127, 127{]} (256
  levels).
\item
  \textbf{INT6}: 6-bit signed integer, implemented by rounding to
  multiples of 4 in the INT8 integer space, yielding effective values in
  \{−124, −120, \ldots, 120, 124\} (\textasciitilde64 levels).
\item
  \textbf{INT4}: 4-bit signed integer, implemented by rounding to
  multiples of 16 in the INT8 integer space, yielding 16 representable
  levels in {[}−128, 112{]}.
\end{itemize}

All quantised modes apply the STE: the forward pass uses the quantised
weight, the backward pass treats the quantisation step as the identity
function.

\subsubsection{3.3 Learning-rate schedule}\label{learning-rate-schedule}

The default optimiser is Muon (Liu et al. 2025) for all 2D matrix
parameters and AdamW for scalar parameters. The Phase 5 D1 ablation
replaces Muon with AdamW for matrix parameters; in that case all
parameters are optimised by AdamW. The LR schedule is
Warmup-Stable-Decay (WSD) (Hu et al. 2024):

\[\text{lr}(t) = \begin{cases}
\text{lr}_{\max} \cdot t / T_{\text{warmup}} & t < T_{\text{warmup}} \\
\text{lr}_{\max} & T_{\text{warmup}} \le t < T - T_{\text{warmdown}} \\
\text{lr}_{\max} \cdot f\!\left(\frac{t - (T - T_{\text{warmdown}})}{T_{\text{warmdown}}}\right) & T - T_{\text{warmdown}} \le t \le T
\end{cases}\]

The warmdown function \(f\) is \textbf{linear}, \(f(u) = 1 - u\), in
Phase 2 and in Phase 5 D1/D2/D4/D5/M2. Phase 5 D3 substitutes a
\textbf{half-cosine} warmdown, \(f(u) = \tfrac{1}{2}(1 + \cos\pi u)\).
All runs use \(T_{\text{warmup}} = 100\) steps and \(T = 9000\) except
D2 which uses \(T \in \{27000, 81000\}\). The four warmdown fractions
used in Phase 2:

{\def\LTcaptype{none} 
\begin{longtable}[]{@{}
  >{\raggedright\arraybackslash}p{(\linewidth - 8\tabcolsep) * \real{0.0889}}
  >{\raggedright\arraybackslash}p{(\linewidth - 8\tabcolsep) * \real{0.1111}}
  >{\raggedright\arraybackslash}p{(\linewidth - 8\tabcolsep) * \real{0.4667}}
  >{\raggedright\arraybackslash}p{(\linewidth - 8\tabcolsep) * \real{0.1667}}
  >{\raggedright\arraybackslash}p{(\linewidth - 8\tabcolsep) * \real{0.1667}}@{}}
\toprule\noalign{}
\begin{minipage}[b]{\linewidth}\raggedright
Tag
\end{minipage} & \begin{minipage}[b]{\linewidth}\raggedright
Fraction
\end{minipage} & \begin{minipage}[b]{\linewidth}\raggedright
Phase 2 (\(T=9000\)) \(T_{\text{warmdown}}\)
\end{minipage} & \begin{minipage}[b]{\linewidth}\raggedright
At \(T=27000\)
\end{minipage} & \begin{minipage}[b]{\linewidth}\raggedright
At \(T=81000\)
\end{minipage} \\
\midrule\noalign{}
\endhead
\bottomrule\noalign{}
\endlastfoot
wd00 & 0\% & 0 & 0 & 0 \\
wd10 & 10\% & 900 & 2700 & 8100 \\
wd33 & 33\% & 2970 & 8910 & 26730 \\
wd50 & 50\% & 4500 & 13500 & 40500 \\
\end{longtable}
}

\subsubsection{3.4 LR magnitude}\label{lr-magnitude}

The Muon reference LR was calibrated on Phase 1 pilots (30M, FP16, wd33,
2000 steps) and locked at \texttt{MATRIX\_LR\ =\ SCALAR\_LR\ =\ 0.0125},
\texttt{TIED\_EMBED\_LR\ =\ 0.0175}. The three LR-magnitude conditions
in Phase 2 scale these together:

{\def\LTcaptype{none} 
\begin{longtable}[]{@{}llll@{}}
\toprule\noalign{}
Tag & MATRIX\_LR & SCALAR\_LR & TIED\_EMBED\_LR \\
\midrule\noalign{}
\endhead
\bottomrule\noalign{}
\endlastfoot
lr05 & 0.00625 & 0.00625 & 0.00875 \\
lr1x & 0.01250 & 0.01250 & 0.01750 \\
lr2x & 0.02500 & 0.02500 & 0.03500 \\
\end{longtable}
}

The lr1x condition is the primary one reported in figures. For Phase 5
D1, an AdamW-specific reference LR was calibrated via a 3-cell pilot
(FP16, wd33, 30M, seed=1337) at MATRIX\_LR ∈ \{0.001, 0.003, 0.0125\}:
0.001 was selected (final val\_bpb 1.2499, vs 1.2687 for 0.003 and
1.3667 for 0.0125). All D1 cells use
\texttt{MATRIX\_LR\ =\ SCALAR\_LR\ =\ 0.001},
\texttt{TIED\_EMBED\_LR\ =\ 0.0014}. The 32 mBPB gap to Muon at
FP16/wd33/30M is reported as part of the D1 result.

\subsubsection{3.5 Phase 2 factorial
design}\label{phase-2-factorial-design}

The Phase 2 grid is fully crossed:
\[3 \text{ bit-widths} \times 4 \text{ warmdowns} \times 3 \text{ LR magnitudes} \times 4 \text{ sizes} \times 5 \text{ seeds} = 720 \text{ runs}.\]
Seeds: \{1337, 42, 0, 7, 2024\}. Each run is 9000 iterations on a single
shard sequence drawn deterministically from the seed; validation BPB is
evaluated at steps 1500, 3000, 4500, 6000, 7500, and 9000, with the
step-9000 value used as the final result.

\subsubsection{3.6 Phase 5 ablation grid}\label{phase-5-ablation-grid}

Phase 5 comprises seven ablations (625 cells total):

{\def\LTcaptype{none} 
\begin{longtable}[]{@{}
  >{\raggedright\arraybackslash}p{(\linewidth - 4\tabcolsep) * \real{0.3846}}
  >{\raggedright\arraybackslash}p{(\linewidth - 4\tabcolsep) * \real{0.2692}}
  >{\raggedright\arraybackslash}p{(\linewidth - 4\tabcolsep) * \real{0.3462}}@{}}
\toprule\noalign{}
\begin{minipage}[b]{\linewidth}\raggedright
Ablation
\end{minipage} & \begin{minipage}[b]{\linewidth}\raggedright
Cells
\end{minipage} & \begin{minipage}[b]{\linewidth}\raggedright
Purpose
\end{minipage} \\
\midrule\noalign{}
\endhead
\bottomrule\noalign{}
\endlastfoot
D1 (AdamW) & 240 & Replicate Phase 2 grid at lr1x with AdamW instead of
Muon \\
D2 (longtrain) & 81 & 30M × \{4 wd\} × \{3 bits\} × \{3 seeds\} × \{27k,
81k iters\} + 9 100M cells at 27k wd33 \\
D3 (cosine) & 60 & wd33 × \{3 bits\} × \{4 sizes\} × \{5 seeds\} with
cosine warmdown \\
D4 (sizes) & 75 & wd33/lr1x × \{3 bits\} × \{5 new sizes\} × \{5
seeds\} \\
D5 (INT4) & 80 & \{4 wd\} × \{4 Phase 2 sizes\} × \{5 seeds\}, all
INT4 \\
D6 (INT4 small) & 80 & \{4 wd\} × \{3M, 5M, 8M, 10M\} × \{5 seeds\}, all
INT4 --- sub-15M extension of D5 \\
M2 (snapshots) & 9 & 30M × \{3 bits\} × \{3 seeds\} at wd33/lr1x, with
weight-to-grid distance logged every 200 steps \\
\end{longtable}
}

Seeds for D2 and M2 are reduced to \{1337, 42, 0\} (3 seeds) for budget;
all other Phase 5 ablations use the full 5-seed set. D2 27k cells use
warmdown lengths scaled to 33\% of \(T\) as listed in §3.3; the same
scaling is used for D2 81k cells. D6 uses two new size configurations
(3M: dim=144, layers=8; 10M: dim=256, layers=12), reusing the D4
configurations for 5M and 8M.

\begin{center}\rule{0.5\linewidth}{0.5pt}\end{center}

\subsection{4. Experiments}\label{experiments}

\subsubsection{4.1 Compute}\label{compute}

All experiments ran on MareNostrum 5 ACC (Barcelona Supercomputing
Center) under EuroHPC AI Factory grant EHPC-AIF-2026PG01-401. Each run
used 4× H100 80GB GPUs with \texttt{grad\_accum\_steps=2} and 32
sequences per micro-batch (\texttt{TRAIN\_BATCH\_TOKENS=65536},
\texttt{TRAIN\_SEQ\_LEN=2048}). Wall-clock per 9000-iteration run scaled
from \textasciitilde3 minutes at 3M to \textasciitilde25 minutes at
100M; the 81k D2 cells took \textasciitilde95 minutes. Total compute
consumed: \textasciitilde580 GPU-hours for Phase 2 and
\textasciitilde1320 GPU-hours for Phase 5. The full
computational-footprint estimate is given in Appendix A.

\subsubsection{4.2 Data and tokenisation}\label{data-and-tokenisation}

Training and evaluation use the FineWeb 10B corpus (Penedo et al. 2024)
tokenised with a SentencePiece BPE tokeniser of vocabulary 8192.
Training shards cover 10B tokens; validation BPB is computed on a
held-out FineWeb validation split (40.5M tokens). Bits-per-byte (BPB) is
the validation metric throughout, computed as
\(\text{BPB} = \text{val\_loss} / (\ln 2 \cdot \text{bytes\_per\_token})\).
We report BPB rather than perplexity to make precision/quantisation gaps
directly interpretable in compression-equivalent terms.

\subsubsection{4.3 Hyperparameters held
constant}\label{hyperparameters-held-constant}

The following are fixed across all 1345 cells (Phase 2 + Phase 5):
\texttt{TRAIN\_BATCH\_TOKENS=65536}, \texttt{TRAIN\_SEQ\_LEN=2048},
\texttt{WARMUP\_STEPS=100}, \texttt{MUON\_MOMENTUM=0.95},
\texttt{MUON\_WEIGHT\_DECAY=0.0}, \texttt{ADAM\_WEIGHT\_DECAY=0.0}, no
EMA, no SWA, no test-time training. The grids vary only the dimensions
specified in §3.5 and §3.6.

\subsubsection{4.4 Statistical
methodology}\label{statistical-methodology}

Each grid cell is summarised by mean BPB across seeds with standard
error of the mean (SEM). Standard errors are 0.4--0.7 mBPB across Phase
2 (5 seeds) and 0.5--1.0 mBPB across D2 and M2 (3 seeds). For penalty
estimates we compute the per-seed paired difference (e.g., INT6 BPB −
FP16 BPB) within each (size, seed) pair and report the mean ± 95\% CI of
those differences. This pairing removes seed-level variance and yields
tighter intervals than unpaired comparisons.

For the prediction analysis (§5.2), the log-linear fit
\texttt{penalty\ =\ a\ +\ b\ ·\ log(size\_M)} is performed on Phase 2
sizes only (15M, 30M, 50M, 100M). The Phase 5 D4 held-out sizes (5M, 8M,
175M, 250M, 350M) are used solely for prediction-interval validation and
are not seen during fitting.

We do not apply multiple-comparison corrections to the
schedule-interaction null result because the design pre-specified one
primary hypothesis (the LR × bit-width interaction at FP16/INT8/INT6)
and all reported model-size × schedule combinations are descriptive
ablations of the same prediction.

\begin{center}\rule{0.5\linewidth}{0.5pt}\end{center}

\subsection{5. Results}\label{results}

We organise results around four questions: (i) does the optimal LR
schedule depend on bit-width for FP16/INT8/INT6 (§5.1)? (ii) how does
the absolute INT6 cost scale with model size, and does the scaling
predict held-out sizes (§5.2)? (iii) where does the null result break,
and how (§5.3)? (iv) is the null result robust to changes in optimiser,
schedule shape, and training length, and is the mechanism what we
initially thought (§5.4--5.6)?

\subsubsection{5.1 The Phase 2 null result on schedule ×
bit-width}\label{the-phase-2-null-result-on-schedule-bit-width}

\begin{figure}
\centering
\includegraphics[width=1\linewidth,height=\textheight,keepaspectratio,alt={Schedule × bit-width interaction at the reference LR (lr1x), Phase 2. Validation BPB as a function of warmdown fraction for FP16, INT8, and INT6 at each model size. Error bars are ±1 SEM across 5 seeds. All three bit-width curves are statistically indistinguishable in shape at every size; all peak at wd33 = 33\% warmdown.}]{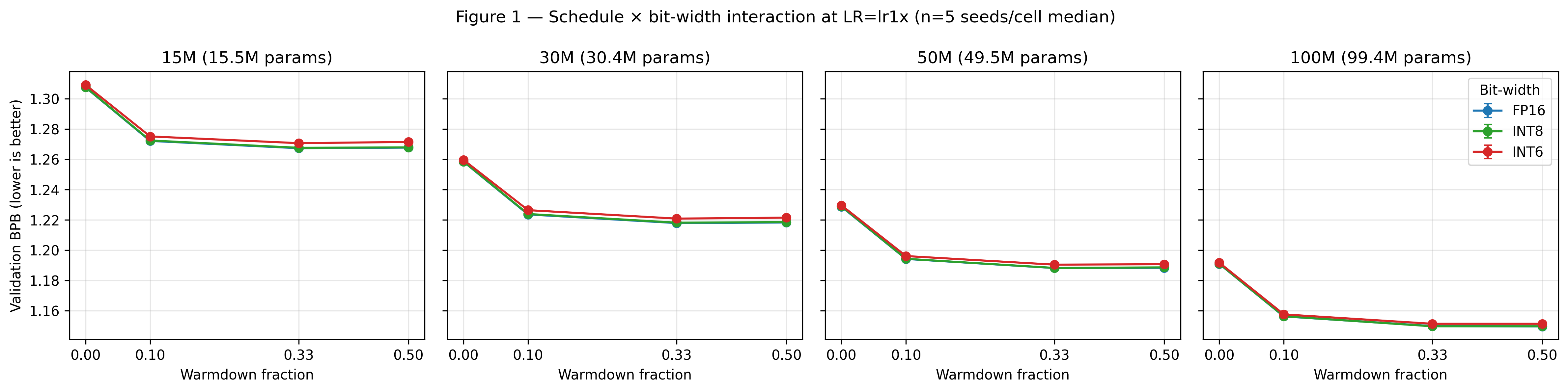}
\caption{\textbf{Schedule × bit-width interaction at the reference LR
(lr1x), Phase 2.} Validation BPB as a function of warmdown fraction for
FP16, INT8, and INT6 at each model size. Error bars are ±1 SEM across 5
seeds. All three bit-width curves are statistically indistinguishable in
shape at every size; all peak at wd33 = 33\% warmdown.}
\end{figure}

The optimal warmdown fraction is identical across all three bit-widths
at every size (Figure 1). The three curves overlap within standard
errors at every (size, warmdown) point, and all peak at wd33. No
reliable interaction term is present at any of the four model sizes
tested.

The warmdown effect itself is large and practically significant. At 30M
parameters, moving from no warmdown (wd00) to the optimal 33\% warmdown
(wd33) recovers approximately 0.040 BPB --- a gap larger than the
difference between INT6 and FP16 at any schedule. This benefit is fully
shared across bit-widths; quantising to INT6 does not change the
schedule that a practitioner should use.

Results at the reduced (lr05) and elevated (lr2x) learning-rate
magnitudes are qualitatively identical: wd33 remains the per-cell
optimum, and the FP16/INT8/INT6 curves remain statistically
indistinguishable in shape. The null result is not an LR-magnitude
artifact.

\textbf{Primary hypothesis (LR × bit-width interaction for
FP16/INT8/INT6): falsified.} The hypothesis that INT6 would benefit from
a longer warmdown relative to higher-precision training is not
supported.

\subsubsection{5.2 INT6 penalty scales log-linearly, and the fit
predicts held-out
sizes}\label{int6-penalty-scales-log-linearly-and-the-fit-predicts-held-out-sizes}

\begin{figure}
\centering
\includegraphics[width=0.85\linewidth,height=\textheight,keepaspectratio,alt={INT6 quantisation cost decreases with model size, and the fit predicts. Paired per-seed penalty (INT6 BPB − FP16 BPB) at the reference condition (lr1x, wd33) as a function of model size. Blue points: Phase 2 measurements (15M, 30M, 50M, 100M) used for the log-linear fit. Red squares: Phase 5 D4 measurements at five held-out sizes (5M, 8M, 175M, 250M, 350M). The fit predicts all five held-out points within the 95\% prediction interval (5/5).}]{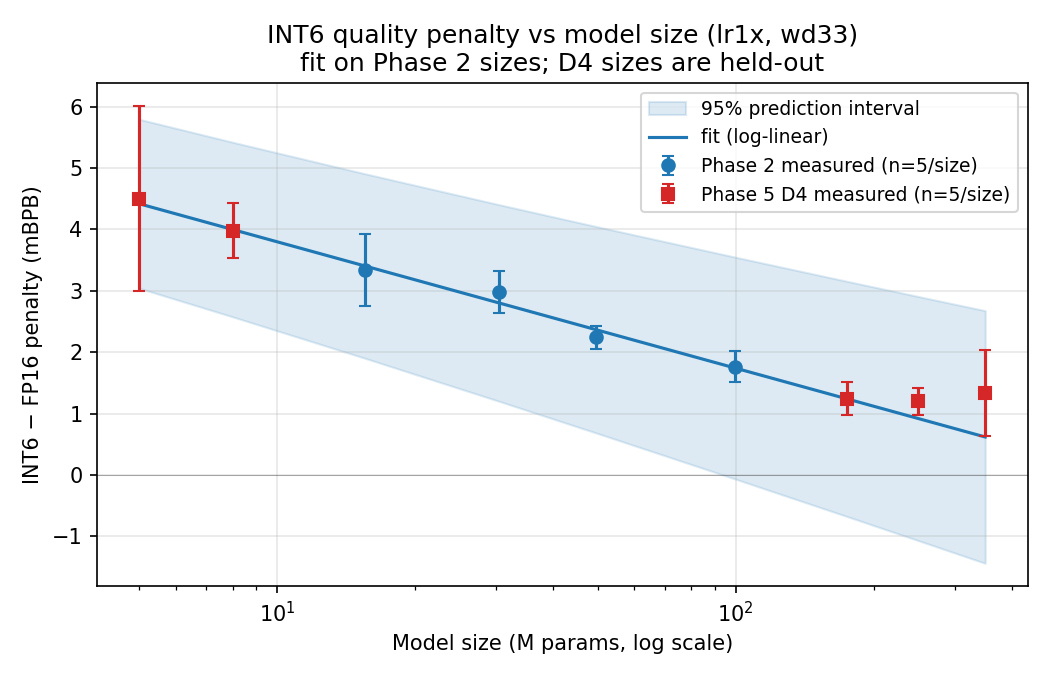}
\caption{\textbf{INT6 quantisation cost decreases with model size, and
the fit predicts.} Paired per-seed penalty (INT6 BPB − FP16 BPB) at the
reference condition (lr1x, wd33) as a function of model size. Blue
points: Phase 2 measurements (15M, 30M, 50M, 100M) used for the
log-linear fit. Red squares: Phase 5 D4 measurements at five held-out
sizes (5M, 8M, 175M, 250M, 350M). The fit predicts all five held-out
points within the 95\% prediction interval (5/5).}
\end{figure}

A clean monotonic trend emerges in the absolute cost of INT6
quantisation (Table 1). At the reference condition (lr1x, wd33), the BPB
gap between INT6 and FP16 decreases with model size from +3.3 mBPB at
15M to +1.8 mBPB at 100M.

\textbf{Table 1.} Validation BPB (mean ± SEM, 5 seeds) at the reference
condition (lr1x, wd33). INT6 penalty is the paired per-seed difference
INT6 BPB − FP16 BPB.

{\def\LTcaptype{none} 
\begin{longtable}[]{@{}
  >{\raggedright\arraybackslash}p{(\linewidth - 8\tabcolsep) * \real{0.0972}}
  >{\raggedright\arraybackslash}p{(\linewidth - 8\tabcolsep) * \real{0.2361}}
  >{\raggedright\arraybackslash}p{(\linewidth - 8\tabcolsep) * \real{0.2361}}
  >{\raggedright\arraybackslash}p{(\linewidth - 8\tabcolsep) * \real{0.2361}}
  >{\raggedright\arraybackslash}p{(\linewidth - 8\tabcolsep) * \real{0.1944}}@{}}
\toprule\noalign{}
\begin{minipage}[b]{\linewidth}\raggedright
Size
\end{minipage} & \begin{minipage}[b]{\linewidth}\raggedright
FP16 BPB
\end{minipage} & \begin{minipage}[b]{\linewidth}\raggedright
INT8 BPB
\end{minipage} & \begin{minipage}[b]{\linewidth}\raggedright
INT6 BPB
\end{minipage} & \begin{minipage}[b]{\linewidth}\raggedright
INT6 penalty
\end{minipage} \\
\midrule\noalign{}
\endhead
\bottomrule\noalign{}
\endlastfoot
15M & 1.2674 ± 0.0007 & 1.2676 ± 0.0005 & 1.2707 ± 0.0006 & +3.3 mBPB \\
30M & 1.2181 ± 0.0007 & 1.2182 ± 0.0007 & 1.2211 ± 0.0006 & +3.0 mBPB \\
50M & 1.1882 ± 0.0004 & 1.1883 ± 0.0004 & 1.1905 ± 0.0004 & +2.2 mBPB \\
100M & 1.1496 ± 0.0005 & 1.1497 ± 0.0005 & 1.1514 ± 0.0006 & +1.8
mBPB \\
\end{longtable}
}

A log-linear fit on Phase 2 sizes yields
\(\text{penalty} = +5.85 - 0.89 \cdot \log(\text{size}_M)\) mBPB (slope
SE = 0.14 mBPB/log-M; residual σ = 0.42 mBPB). The fit's predictions at
the five Phase 5 D4 held-out sizes are reported in Table 2.

\textbf{Table 2.} Predicted vs measured INT6 penalty at held-out sizes
(Phase 5 D4). The fit is computed on Phase 2 sizes only; D4 points are
unseen during fitting. All five held-out points fall within their 95\%
prediction intervals.

{\def\LTcaptype{none} 
\begin{longtable}[]{@{}
  >{\raggedright\arraybackslash}p{(\linewidth - 10\tabcolsep) * \real{0.0814}}
  >{\raggedright\arraybackslash}p{(\linewidth - 10\tabcolsep) * \real{0.2093}}
  >{\raggedright\arraybackslash}p{(\linewidth - 10\tabcolsep) * \real{0.2093}}
  >{\raggedright\arraybackslash}p{(\linewidth - 10\tabcolsep) * \real{0.1163}}
  >{\raggedright\arraybackslash}p{(\linewidth - 10\tabcolsep) * \real{0.2442}}
  >{\raggedright\arraybackslash}p{(\linewidth - 10\tabcolsep) * \real{0.1395}}@{}}
\toprule\noalign{}
\begin{minipage}[b]{\linewidth}\raggedright
Size
\end{minipage} & \begin{minipage}[b]{\linewidth}\raggedright
Predicted (mBPB)
\end{minipage} & \begin{minipage}[b]{\linewidth}\raggedright
Measured (mBPB)
\end{minipage} & \begin{minipage}[b]{\linewidth}\raggedright
Residual
\end{minipage} & \begin{minipage}[b]{\linewidth}\raggedright
95\% PI half-width
\end{minipage} & \begin{minipage}[b]{\linewidth}\raggedright
Inside PI?
\end{minipage} \\
\midrule\noalign{}
\endhead
\bottomrule\noalign{}
\endlastfoot
5M & +4.42 & +4.50 & +0.08 & ±1.38 & ✓ \\
8M & +4.00 & +3.98 & −0.02 & ±1.42 & ✓ \\
175M & +1.24 & +1.24 & +0.00 & ±1.92 & ✓ \\
250M & +0.92 & +1.20 & +0.28 & ±1.99 & ✓ \\
350M & +0.62 & +1.34 & +0.72 & ±2.06 & ✓ \\
\end{longtable}
}

The fit extrapolates correctly across nearly two orders of magnitude (5M
→ 350M, \textasciitilde70× range). One sub-pattern is worth flagging:
residuals at 175M, 250M, and 350M are all positive (measured
\textgreater{} predicted). The slope may be flattening at scale --- the
INT6 penalty may not continue to decrease as fast as the log-linear fit
predicts. The 95\% PI covers all five points, so we do not reject the
log-linear form; but the systematic positive bias in the largest three
points is a candidate for follow-up.

\subsubsection{5.3 Where the null result breaks: INT4 below
50M}\label{where-the-null-result-breaks-int4-below-50m}

\begin{figure}
\centering
\includegraphics[width=1\linewidth,height=\textheight,keepaspectratio,alt={INT4 schedule sensitivity and the precision boundary, across the 3M--100M size range. Panel (a): INT4 validation BPB vs warmdown fraction, one line per model size, combining the 80-cell Phase 5 D5 grid (15M--100M) with the 80-cell Phase 5 D6 extension (3M--10M). Error bars are 95\% CI across 5 seeds. The wd00 → wd10 step is the dominant feature at every size; the schedule curves are nearly flat from wd10 onwards at smaller sizes. Panel (b): paired wd10 − wd33 penalty (mBPB) vs model size with 95\% CI band. Below 50M the band crosses zero --- the schedule preference is noise-dominated. At 50M and 100M the penalty is decisively positive (z \textgreater{} 10), and wd33 is the unambiguous optimum. The transition between regimes sits between 30M and 50M.}]{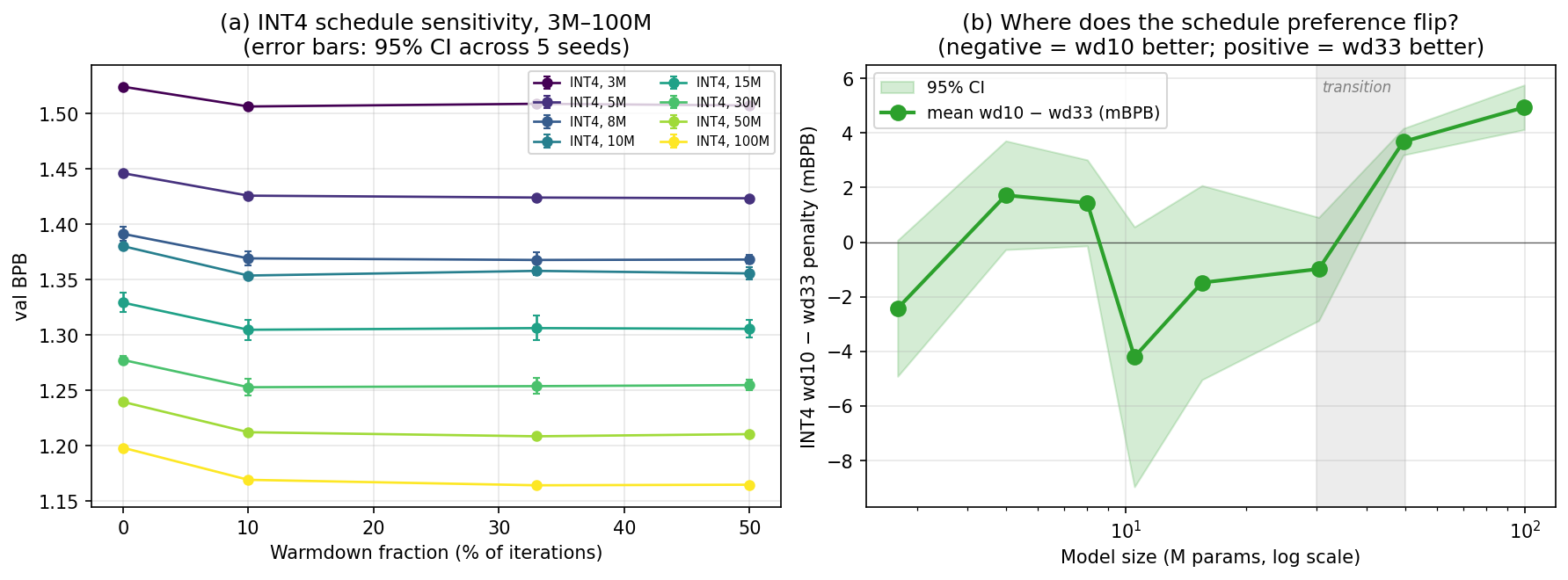}
\caption{\textbf{INT4 schedule sensitivity and the precision boundary,
across the 3M--100M size range.} Panel (a): INT4 validation BPB vs
warmdown fraction, one line per model size, combining the 80-cell Phase
5 D5 grid (15M--100M) with the 80-cell Phase 5 D6 extension (3M--10M).
Error bars are 95\% CI across 5 seeds. The wd00 → wd10 step is the
dominant feature at every size; the schedule curves are nearly flat from
wd10 onwards at smaller sizes. Panel (b): paired wd10 − wd33 penalty
(mBPB) vs model size with 95\% CI band. Below 50M the band crosses zero
--- the schedule preference is noise-dominated. At 50M and 100M the
penalty is decisively positive (z \textgreater{} 10), and wd33 is the
unambiguous optimum. The transition between regimes sits between 30M and
50M.}
\end{figure}

Phase 5 D5 (80 cells, 15M--100M) and Phase 5 D6 (80 cells, 3M--10M)
together provide a full INT4 sweep over warmdown × size with 5 seeds per
cell. The combined picture is more nuanced than a clean wd10 ↔ wd33
boundary.

\textbf{Below 50M, the schedule preference is noise-dominated.} Table 3
reports the paired per-seed penalty (INT4 wd10 BPB − INT4 wd33 BPB) at
every size. The point estimate flips sign with size: 3M, 10M, 15M, and
30M lean wd10 (negative penalty, but with \textbar z\textbar{}
\textless{} 2); 5M and 8M lean wd33 by a similar margin (positive, but
again \textbar z\textbar{} \textless{} 2). No size below 50M shows a
statistically significant preference at the conventional 1.96·SE
threshold. The seed-level vote (how many of 5 seeds prefer wd10 over
wd33 within their seed) is correspondingly mixed: 4/5, 2/5, 2/5, 4/5,
3/5, 4/5 at 3M, 5M, 8M, 10M, 15M, 30M respectively. The wd-shift ``to
wd10'' that a quick scan of mean BPB at 15M/30M suggests is real in
direction but small enough to live within the seed-level noise floor.

\textbf{At 50M and above, wd33 is decisively optimal.} The penalty jumps
from a noisy −1 mBPB at 30M to a strongly significant +3.7 mBPB at 50M
(z ≈ 15, p ≪ 0.001), then to +4.9 mBPB at 100M (z ≈ 12). All 10/10 seed
comparisons at 50M and 100M prefer wd33. This is a sharp transition, not
a gradual one.

\textbf{Table 3.} Paired INT4 schedule penalty (wd10 BPB − wd33 BPB) at
lr1x across the 3M--100M range. \emph{n} = 5 seeds per size; SE is
across paired-seed differences. Significance threshold:
\textbar z\textbar{} ≥ 1.96. Below 50M, no individual size is
significant; at 50M and above, both are significant by a large margin.

{\def\LTcaptype{none} 
\begin{longtable}[]{@{}
  >{\raggedright\arraybackslash}p{(\linewidth - 10\tabcolsep) * \real{0.0824}}
  >{\raggedleft\arraybackslash}p{(\linewidth - 10\tabcolsep) * \real{0.2118}}
  >{\raggedleft\arraybackslash}p{(\linewidth - 10\tabcolsep) * \real{0.1294}}
  >{\raggedleft\arraybackslash}p{(\linewidth - 10\tabcolsep) * \real{0.0941}}
  >{\raggedleft\arraybackslash}p{(\linewidth - 10\tabcolsep) * \real{0.2706}}
  >{\raggedright\arraybackslash}p{(\linewidth - 10\tabcolsep) * \real{0.2118}}@{}}
\toprule\noalign{}
\begin{minipage}[b]{\linewidth}\raggedright
Size
\end{minipage} & \begin{minipage}[b]{\linewidth}\raggedleft
mean wd10 − wd33
\end{minipage} & \begin{minipage}[b]{\linewidth}\raggedleft
SE (mBPB)
\end{minipage} & \begin{minipage}[b]{\linewidth}\raggedleft
z
\end{minipage} & \begin{minipage}[b]{\linewidth}\raggedleft
seeds preferring wd10
\end{minipage} & \begin{minipage}[b]{\linewidth}\raggedright
verdict
\end{minipage} \\
\midrule\noalign{}
\endhead
\bottomrule\noalign{}
\endlastfoot
3M & −2.42 mBPB & 1.27 & −1.90 & 4 / 5 & noise \\
5M & +1.72 mBPB & 1.02 & +1.69 & 2 / 5 & noise \\
8M & +1.44 mBPB & 0.80 & +1.79 & 2 / 5 & noise \\
10M & −4.20 mBPB & 2.43 & −1.73 & 4 / 5 & noise \\
15M & −1.48 mBPB & 1.82 & −0.82 & 3 / 5 & noise \\
30M & −0.98 mBPB & 0.96 & −1.02 & 4 / 5 & noise \\
\textbf{50M} & \textbf{+3.68 mBPB} & \textbf{0.25} & \textbf{+14.85} &
\textbf{0 / 5} & \textbf{wd33 (sig.)} \\
\textbf{100M} & \textbf{+4.94 mBPB} & \textbf{0.42} & \textbf{+11.86} &
\textbf{0 / 5} & \textbf{wd33 (sig.)} \\
\end{longtable}
}

The right framing for the schedule × INT4 interaction is therefore not
``wd10 wins at small models'' but rather \textbf{``below 50M, the choice
between wd10, wd33, and wd50 is in the noise; at 50M and above, wd33
wins decisively''}. The sharp transition between these regimes sits
between 30M and 50M.

This is a more conservative claim than a naïve reading of the per-size
means would license, and we flag it as a correction: an earlier draft of
this paper, working from D5 alone, reported that ``all five seeds at 15M
and 30M prefer wd10 over wd33'' as evidence ruling out noise.
Re-examination shows the actual per-seed splits at those sizes are 3/5
and 4/5 respectively --- not unanimous, and not significant under the
paired test.

The INT4 quality penalty itself (versus FP16) is an order of magnitude
larger than the INT6 penalty (Table 4) and decreases with model size,
much like the INT6 penalty does. The largest single drop in the INT4
penalty across the size range is the 30M → 50M step (from +35.9 to +20.4
mBPB), which sits at the same boundary as the schedule-preference
transition. Both the schedule sensitivity and the quality penalty are
roughly co-located in size, consistent with the interpretation that 50M
is the scale at which INT4-quantisation noise has fallen below the
threshold that perturbs the optimiser's late-stage dynamics.

\textbf{Table 4.} INT4 quality penalty at wd33/lr1x compared to INT6,
all relative to FP16. Sizes 15M--100M from Phase 2 + D5; no FP16
baselines were collected for 3M/10M so those rows are omitted.

{\def\LTcaptype{none} 
\begin{longtable}[]{@{}
  >{\raggedright\arraybackslash}p{(\linewidth - 8\tabcolsep) * \real{0.1000}}
  >{\raggedright\arraybackslash}p{(\linewidth - 8\tabcolsep) * \real{0.1429}}
  >{\raggedright\arraybackslash}p{(\linewidth - 8\tabcolsep) * \real{0.2000}}
  >{\raggedright\arraybackslash}p{(\linewidth - 8\tabcolsep) * \real{0.3000}}
  >{\raggedright\arraybackslash}p{(\linewidth - 8\tabcolsep) * \real{0.2571}}@{}}
\toprule\noalign{}
\begin{minipage}[b]{\linewidth}\raggedright
Size
\end{minipage} & \begin{minipage}[b]{\linewidth}\raggedright
FP16 BPB
\end{minipage} & \begin{minipage}[b]{\linewidth}\raggedright
INT6 penalty
\end{minipage} & \begin{minipage}[b]{\linewidth}\raggedright
INT4 penalty (wd33)
\end{minipage} & \begin{minipage}[b]{\linewidth}\raggedright
INT4/INT6 ratio
\end{minipage} \\
\midrule\noalign{}
\endhead
\bottomrule\noalign{}
\endlastfoot
15M & 1.2674 & +3.3 mBPB & +39.0 mBPB & 11.7× \\
30M & 1.2181 & +3.0 mBPB & +35.9 mBPB & 12.0× \\
50M & 1.1882 & +2.2 mBPB & +20.4 mBPB & 9.1× \\
100M & 1.1496 & +1.8 mBPB & +14.8 mBPB & 8.4× \\
\end{longtable}
}

\subsubsection{5.4 The null result is robust to optimiser, schedule
shape, and training
length}\label{the-null-result-is-robust-to-optimiser-schedule-shape-and-training-length}

\begin{figure}
\centering
\includegraphics[width=1\linewidth,height=\textheight,keepaspectratio,alt={Robustness of the null result. (a) D1: AdamW at 9k iters @ 30M --- wd33 remains optimal; FP16/INT8/INT6 curves overlap. (b) D2: Muon at 81k iters @ 30M --- wd33 remains optimal at 9× longer training; INT6 sits slightly above FP16/INT8 (the growing penalty, §5.5). (c) D3: linear-WSD vs cosine at wd33 / 30M --- cosine is slightly worse than linear (\textasciitilde+2 mBPB) but the relative bit-width pattern is identical.}]{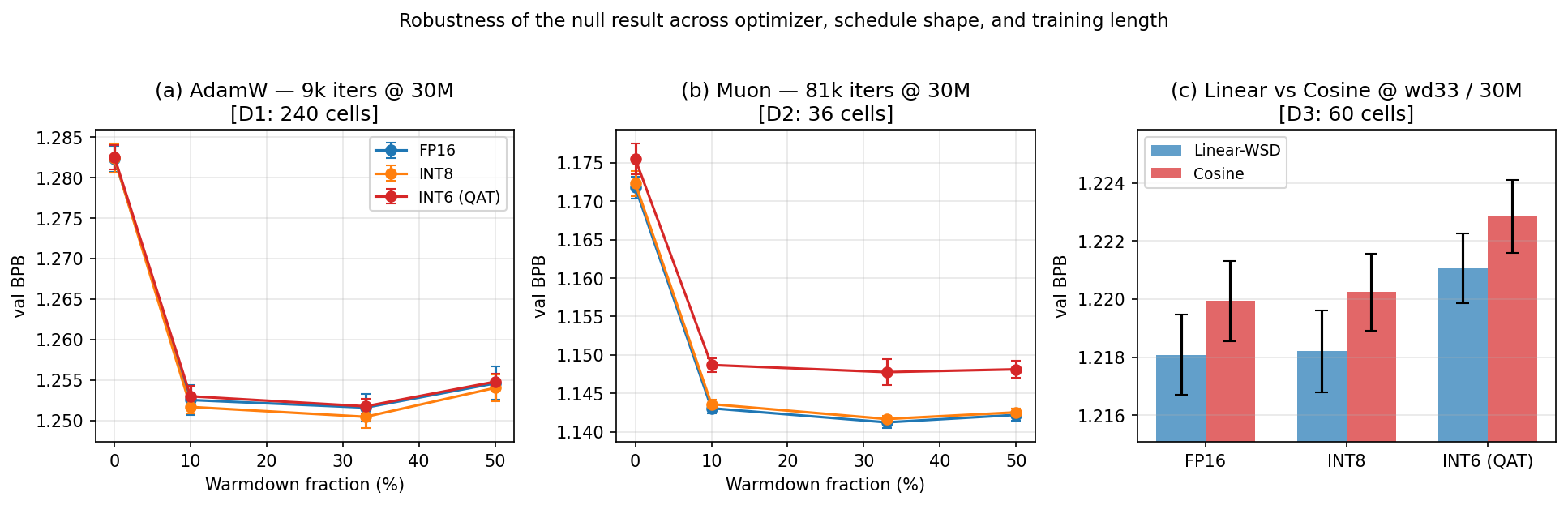}
\caption{\textbf{Robustness of the null result.} (a) D1: AdamW at 9k
iters @ 30M --- wd33 remains optimal; FP16/INT8/INT6 curves overlap. (b)
D2: Muon at 81k iters @ 30M --- wd33 remains optimal at 9× longer
training; INT6 sits slightly above FP16/INT8 (the growing penalty,
§5.5). (c) D3: linear-WSD vs cosine at wd33 / 30M --- cosine is slightly
worse than linear (\textasciitilde+2 mBPB) but the relative bit-width
pattern is identical.}
\end{figure}

Three independent perturbations of the Phase 2 setup leave the
bit-width-agnostic null result intact:

\textbf{D1 --- AdamW (240 cells).} Replicating the Phase 2 grid at lr1x
with AdamW (matrix LR 0.001 per the calibration in §3.4) instead of
Muon, wd33 is optimal at every (bit-width, size) combination. The
warmdown effect (wd33 vs wd00) is +27 to +34 mBPB, the same magnitude as
Muon. AdamW underperforms Muon by a consistent 33--42 mBPB across all
sizes; the \emph{interaction} with bit-width is identical under both
optimisers, even though the absolute level differs.

\textbf{D2 --- longer training (81 cells).} At 27k and 81k iterations on
the 30M model, wd33 remains optimal for all bit-widths. At 81k iters,
the wd00 → wd33 gap is +28 to +31 mBPB; at 9k iters in Phase 2, the same
gap was +27 to +33 mBPB. The warmdown benefit is conserved across the 9×
iteration range, and so is its bit-width agnosticism.

\textbf{D3 --- cosine schedule (60 cells).} Replacing the linear
warmdown with a half-cosine produces a small but statistically
significant absolute penalty (\textasciitilde+2 mBPB at 50M and 100M;
8/12 cells exceed 1.96·SE). However, the \emph{relative} pattern across
bit-widths is unchanged: cosine FP16 and cosine INT6 at 100M differ by
17 mBPB; linear FP16 and linear INT6 at 100M differ by 18 mBPB. Schedule
shape matters slightly; schedule × bit-width interaction is null under
both shapes.

The combined message of D1 + D2 + D3: the bit-width-agnosticism of the
schedule, observed in Phase 2 for Muon at 9k iterations under a linear
warmdown, transfers to AdamW, to cosine warmdown, and to up to 9× longer
training. This is a substantially stronger statement than the Phase 2
null result alone.

\subsubsection{5.5 INT6 penalty grows with training
length}\label{int6-penalty-grows-with-training-length}

\begin{figure}
\centering
\includegraphics[width=1\linewidth,height=\textheight,keepaspectratio,alt={INT6 penalty as a function of training length. Panel (a): D2 wd × bit-width curves at 81k iterations on 30M. wd33 is still optimal for all three bit-widths; INT6 sits slightly above FP16/INT8 (the growing penalty). Panel (b): INT6 − FP16 penalty at 30M wd33 lr1x as a function of training iterations (9k from Phase 2; 27k, 81k from Phase 5 D2). The penalty grows roughly linearly with log iterations, from +3.0 mBPB at 9k to +6.5 mBPB at 81k.}]{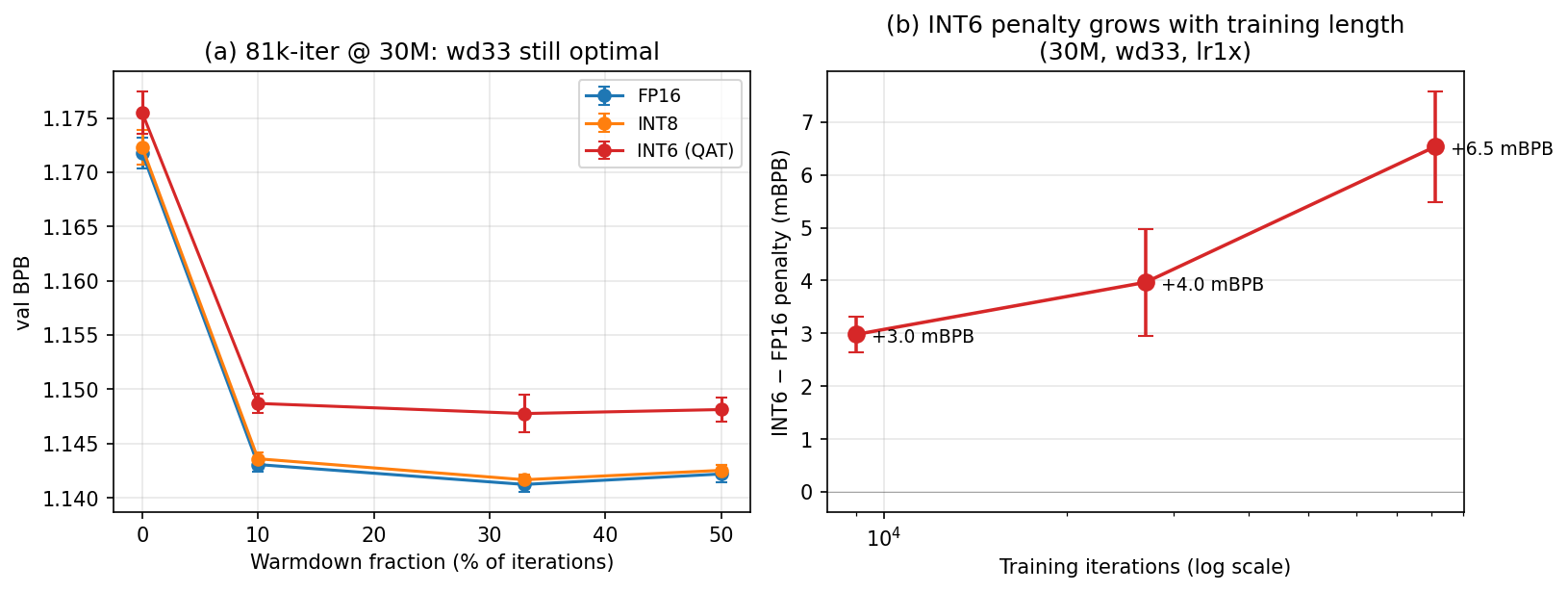}
\caption{\textbf{INT6 penalty as a function of training length.} Panel
(a): D2 wd × bit-width curves at 81k iterations on 30M. wd33 is still
optimal for all three bit-widths; INT6 sits slightly above FP16/INT8
(the growing penalty). Panel (b): INT6 − FP16 penalty at 30M wd33 lr1x
as a function of training iterations (9k from Phase 2; 27k, 81k from
Phase 5 D2). The penalty grows roughly linearly with log iterations,
from +3.0 mBPB at 9k to +6.5 mBPB at 81k.}
\end{figure}

While the \emph{schedule} × bit-width interaction remains null at longer
training, the \emph{absolute} INT6 cost does not. At 30M wd33 lr1x, the
INT6 − FP16 penalty grows from +3.0 mBPB at 9k iterations to +4.0 mBPB
at 27k and +6.5 mBPB at 81k (Figure 5b). The penalty roughly doubles
across the 9× iteration range.

This complements the size-scaling finding from §5.2: the INT6 penalty
\emph{shrinks} with model size and \emph{grows} with training length.
Both trends operate independently within our grid --- at fixed size the
iteration trend is monotonic, and at fixed iteration count the size
trend is monotonic. Combined, the two trends imply that the INT6 cost is
best characterised as a function of both (size, training compute), not
either alone. Dremov et al. (2025) make a related compute-optimal
observation at 86M--2.2B; our 30M result is qualitatively consistent
with theirs at much smaller scale.

The practical implication is mixed. A practitioner training for the
standard 9k-iteration regime can ignore the INT6 cost almost entirely at
100M+. A practitioner training for 81k iterations (or longer) at 30M
should expect a larger penalty (\textasciitilde6.5 mBPB) that may or may
not be worth the deployment savings, depending on the application.

\subsubsection{5.6 Mechanism: it is not
grid-snapping}\label{mechanism-it-is-not-grid-snapping}

\begin{figure}
\centering
\includegraphics[width=0.85\linewidth,height=\textheight,keepaspectratio,alt={Weight-to-INT6-grid distance over training (Phase 5 M2). Mean RMS distance from each 2D linear weight to its INT6-quantised counterpart, averaged across all parameters and across 3 seeds, at 30M wd33 lr1x. FP16 (blue), INT8 (orange), and INT6-QAT (red) all sit at essentially the same distance from the INT6 grid throughout training --- pre-warmdown ratio INT6/FP16 ≈ 1.04. The bit-width-agnosticism of the schedule is therefore not mediated by INT6-QAT weights converging to the grid before warmdown begins.}]{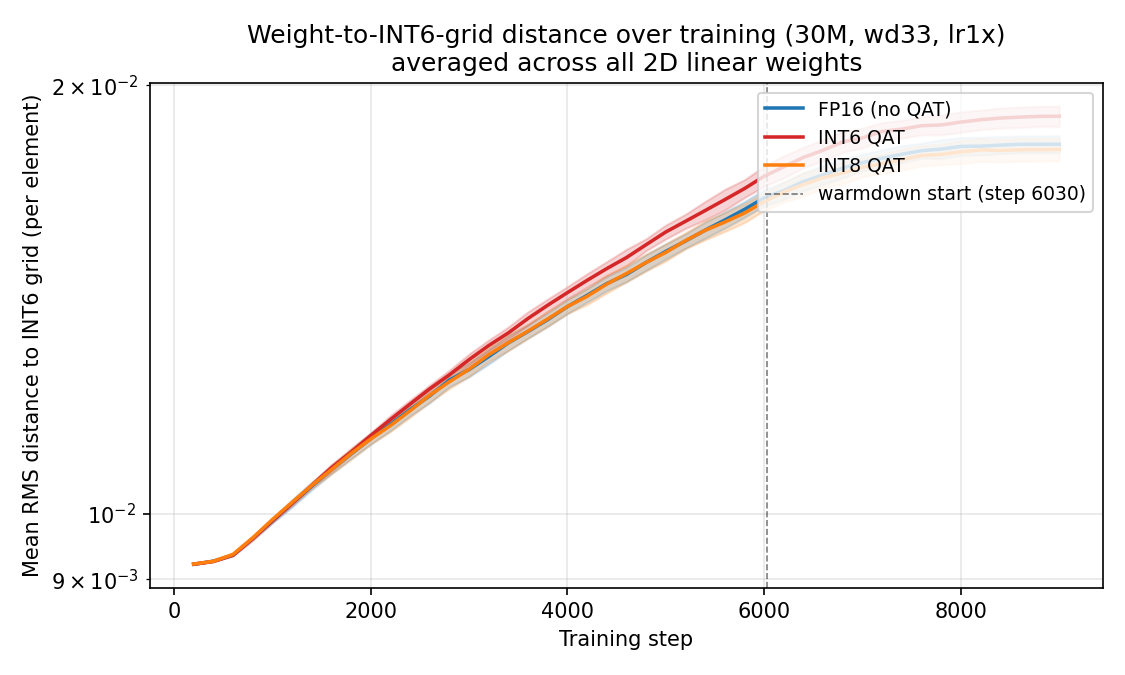}
\caption{\textbf{Weight-to-INT6-grid distance over training (Phase 5
M2).} Mean RMS distance from each 2D linear weight to its INT6-quantised
counterpart, averaged across all parameters and across 3 seeds, at 30M
wd33 lr1x. FP16 (blue), INT8 (orange), and INT6-QAT (red) all sit at
essentially the same distance from the INT6 grid throughout training ---
pre-warmdown ratio INT6/FP16 ≈ 1.04. The bit-width-agnosticism of the
schedule is therefore not mediated by INT6-QAT weights converging to the
grid before warmdown begins.}
\end{figure}

We initially hypothesised that the schedule null result for
FP16/INT8/INT6 would be explained by rapid early convergence of INT6-QAT
weights to the quantisation grid, leaving no destabilisation for the
warmdown phase to consolidate. This hypothesis is straightforwardly
testable: log each parameter's RMS distance to the INT6 grid every 200
steps, then compare the trajectory across bit-widths. We ran 9 such
snapshot runs (FP16, INT8, INT6 × 3 seeds) at 30M wd33 lr1x.

The hypothesis is falsified. Figure 6 shows that pre-warmdown (step
6000, just before the warmdown begins), the mean weight-to-INT6-grid
distance is 0.0167 for FP16, 0.0166 for INT8, and 0.0173 for INT6 ---
all within 4\% of each other. INT6-QAT weights are \emph{not} closer to
the grid than FP16 weights are; they sit at essentially the same
distance. By the end of training (step 9000), the relative ordering and
magnitudes are unchanged.

This rules out the simplest mechanistic explanation. The
bit-width-agnosticism of the optimal warmdown for FP16/INT8/INT6 is not
driven by INT6-QAT models snapping to the grid early. The actual
mechanism is open. Candidates that remain compatible with both the
schedule null result for FP16/INT8/INT6 \emph{and} the INT4 regime
structure (noise-dominated below 50M, decisive wd33 above) include: (i)
the STE producing essentially full-precision gradient signal so
optimisation dynamics are similar across bit-widths until quantisation
noise dominates training noise (which would happen at INT4 first); (ii)
effective regularisation from STE oscillation (Nagel et al. 2022) being
approximately bit-width-invariant in the regime where the quantisation
step is small relative to per-parameter variance, and stronger at INT4
where the step grows; (iii) curvature properties of the loss landscape
rather than weight-space distance to the grid. Distinguishing among
these is left to future work.

\begin{center}\rule{0.5\linewidth}{0.5pt}\end{center}

\subsection{6. Discussion}\label{discussion}

\subsubsection{6.1 The bounded null result and the location of the
boundary}\label{the-bounded-null-result-and-the-location-of-the-boundary}

The first lesson of this paper is the resilience of the FP16/INT8/INT6
null result: it survives changing the optimiser (Muon → AdamW), the
schedule shape (linear → cosine), the training length (9k → 27k → 81k),
and the model size range (down to 5M, up to 350M). Practitioners
deploying weight-only QAT at INT8 or INT6 below 100M can tune the
schedule once at FP16 and reuse it; nothing in our seven ablations
contradicts that recommendation.

The second lesson is the structure of the INT4 boundary. Combining D5
(15M--100M) with D6 (3M--10M) gives an INT4 schedule sweep at eight
sizes spanning more than a 30× range. Two regimes emerge sharply: above
50M, wd33 is decisively optimal (paired z ≈ 12--15, 10 of 10 seed
comparisons prefer wd33 at 50M and 100M); below 50M, no individual size
has a statistically significant schedule preference, and the per-size
mean penalty oscillates within seed-level noise (Table 3, Figure 3b).
The right framing for the schedule × INT4 interaction is therefore not
``INT4 at small models prefers wd10'' --- a claim that would require
seed-level consistency we do not observe --- but rather ``the schedule
preference is decisive at 50M and above, and noise-dominated below
50M''. The transition between regimes sits between 30M and 50M and is
co-located with the largest single drop in the INT4 quality penalty
(Table 4). We flag the more conservative claim explicitly: an earlier
draft of this paper reported that all five seeds at 15M and 30M prefer
wd10 over wd33; re-examination shows the actual per-seed splits are 3/5
and 4/5, not unanimous, and the paired tests at those sizes do not reach
significance.

\subsubsection{6.2 Why might INT4 below 50M look different at
all?}\label{why-might-int4-below-50m-look-different-at-all}

Three observations constrain a candidate explanation. (a) INT4
weight-to-grid distance is much larger than INT6 (the quantisation step
is 4× wider). (b) The small-model regime has the highest per-parameter
quantisation noise, since the FP16 weight distribution narrows as model
size grows. (c) The recovery of decisively significant wd33-optimality
by 50M happens at the same scale where the INT4 quality penalty drops
most rapidly (Table 4: 30M → 50M drops the INT4 penalty from 36 to 20
mBPB, the largest single drop in the size sweep). Together these suggest
that below 50M, INT4-induced quantisation noise has become comparable to
other sources of training noise, and the schedule choice no longer
cleanly identifies a single optimum --- different seeds end up
preferring different warmdowns. By 50M, the per-parameter quantisation
noise has fallen enough that the standard wd33 schedule is again
unambiguously preferred. We emphasise that this is a story about
\emph{which regime the optimiser is in}, not about a specific
alternative warmdown winning below 50M.

This is consistent with the M2 finding (§5.6) that the
bit-width-agnostic schedule for FP16/INT8/INT6 is \emph{not} about
grid-distance: those bit-widths all produce weights at the same grid
distance because the quantisation noise at INT6 is still small relative
to other noise sources. INT4 finally pushes quantisation noise above
that threshold, but only at small models. The story is one of
\emph{noise relative magnitude}, not \emph{grid distance per se}.

\subsubsection{6.3 The mechanism is not
grid-snapping}\label{the-mechanism-is-not-grid-snapping}

We initially expected that the bit-width-agnostic schedule for
FP16/INT8/INT6 could be explained by a mechanism like ``INT6-QAT weights
snap to the quantisation grid early; by the time the warmdown begins,
there is nothing left for the schedule to consolidate.'' The M2
weight-to-grid trajectories rule this out cleanly. Pre-warmdown
weight-to-grid distance is essentially identical (within
\textasciitilde4\%) for FP16, INT8, and INT6 training runs --- measured
against the \emph{same} INT6 grid for all three. INT6-QAT models are not
closer to the grid than FP16 models. Whatever the actual mechanism is,
it is not grid-snapping in any naive sense.

This is a useful negative result. It eliminates one obvious candidate
and constrains the search for the actual mechanism to explanations that
operate at the gradient or loss-landscape level rather than at the
weight-space level. §5.6 lists three candidates that survive M2;
distinguishing among them would require either second-derivative probes
of the loss landscape or controlled experiments at INT4 (where the noise
scale changes) to test the relative-noise hypothesis directly.

\subsubsection{6.4 INT6 penalty: scaling in both
directions}\label{int6-penalty-scaling-in-both-directions}

The INT6 penalty (vs FP16) at wd33/lr1x has two clean monotonic
scalings. With size at fixed iteration count, the penalty shrinks
roughly log-linearly (slope ≈ −0.9 mBPB/log-M; §5.2, Figure 2). With
iteration count at fixed size, the penalty grows roughly log-linearly (≈
+1.6 mBPB per doubling of iterations at 30M; §5.5, Figure 5b). The two
trends do not collapse to a single compute axis within our grid: 30M at
27k iterations and 100M at 9k iterations are within \textasciitilde10\%
in total FLOPs but their INT6 penalties differ substantially (+4.0 mBPB
vs +1.8 mBPB). Size and training length contribute independently. A
joint (size, iterations) sweep that would map the full compute-axis
scaling is left to future work.

\subsubsection{6.5 Linear vs cosine
warmdown}\label{linear-vs-cosine-warmdown}

D3 finds that linear warmdown outperforms cosine warmdown by
\textasciitilde2 mBPB at wd33 (50M and 100M; statistically significant).
The effect is small and orthogonal to the bit-width-agnosticism. Why
linear wins by this margin is not obvious; linear keeps the LR higher
for most of the warmdown phase, which may produce a more useful
weight-space trajectory than the cosine's softer decay. We did not
investigate other warmdown shapes; cosine-to-floor (the variant used by
Apple (Dremov et al. 2025)) is a candidate that may close the gap.

\subsubsection{6.6 Practical
recommendations}\label{practical-recommendations}

From the combined Phase 2 + Phase 5 results, we can issue four narrow
recommendations for sub-100M from-initialisation QAT:

\begin{enumerate}
\def\labelenumi{\arabic{enumi}.}
\tightlist
\item
  \textbf{Use wd33 linear warmdown.} For FP16, INT8, INT6 at any size
  15M--350M, this is optimal within our grid. The schedule transfers
  across bit-widths without retuning. For INT4 at 50M and above, the
  same recommendation applies and is in fact more strongly supported (z
  ≈ 12--15).
\item
  \textbf{For INT4 below 50M, the schedule choice is in the noise.}
  Across six tested sizes from 3M to 30M, no individual size has a
  statistically significant preference among wd10, wd33, and wd50. wd33
  remains a defensible default; wd10 is approximately equivalent within
  seed-level error. Practitioners should not invest tuning effort here.
\item
  \textbf{AdamW is not a drop-in for Muon at this scale.} AdamW
  underperforms Muon by 33--42 mBPB across all sizes at our reference
  setup, even after LR recalibration. The schedule × bit-width null
  result is preserved, but the absolute level is worse. Use Muon if
  available.
\item
  \textbf{Expect the INT6 cost to grow with training length.} At 9k
  iterations, INT6 vs FP16 is \textasciitilde3 mBPB at 30M. At 81k
  iterations, it is \textasciitilde6.5 mBPB. Practitioners running far
  past compute-optimal should budget for this.
\end{enumerate}

\subsubsection{6.7 Comparison to related
work}\label{comparison-to-related-work}

The closest related work in setup is Nielsen et al. (2025), which
studies schedule transitions during continual QAT for sub-50M ternary
models. Their focus on continual QAT (post-training schedule
modification, with an FP warm-start) is orthogonal to our
from-initialisation QAT setup; they find that the schedule at the QAT
transition point matters significantly, while we find that for
from-initialisation QAT the schedule × bit-width effect is null for
FP16/INT8/INT6 and emerges at INT4 only below \textasciitilde50M. The
two results are not contradictory --- they reflect different training
paradigms.

Dremov et al. (2025) operate at 86M--2.2B and find that the optimal peak
LR decreases with bit-width at scale. Our 100M result sits at the
boundary of their range; the interaction they observe is absent in our
null result at FP16/INT8/INT6 and is replaced, in our INT4 sweep, by a
sharp regime transition at 50M rather than a smooth bit-width-dependent
shift. One interpretation: the interaction Dremov et al.~observe is the
large-model analogue of our INT4 noise-dominated regime --- both reflect
regimes where quantisation noise has become large relative to other
training noise, even though the macroscopic signatures differ (a smooth
LR shift in their setting vs a discrete regime transition in ours).
Verifying this would require either replicating their compute-optimal
protocol at INT4 at scale or replicating our INT4-small-model protocol
at lower precisions still.

\subsubsection{6.8 Limitations}\label{limitations}

This study is restricted to weight-only quantisation (W*A16); activation
quantisation (W6A6 etc.) is not covered. All experiments use a single
tokeniser (SP8192) and a single dataset (FineWeb). The architecture is
decoder-only with tied embeddings, U-net skip connections, GQA, and Muon
optimiser; results may not transfer to encoder or encoder-decoder
architectures, and Phase 5 D1 partially addresses the AdamW question for
matrix parameters but does not test alternative scalar optimisers.
Quantisation is applied uniformly from step 0; gradual-quantisation
curricula (Morreale et al. 2025) and continual-QAT setups (Nielsen et
al. 2025) may produce different schedule-sensitivity profiles. The INT4
boundary is now characterised across 3M--100M (eight sizes) and shows a
sharp regime transition at 30M → 50M, but we did not test below 3M; we
do not know whether the noise-dominated regime extends arbitrarily far
down or whether a different schedule emerges at sub-3M scales. M2 is run
only at one cell (30M wd33 lr1x) and one mechanism candidate (grid
distance); other candidate mechanisms remain untested.

\subsubsection{6.9 Computational
footprint}\label{computational-footprint}

Following the systematic-reporting framework of Henderson et al. (2020),
building on the energy-reporting tradition of Strubell et al. (2019) and
Lacoste et al. (2019), we estimate the energy, carbon, and
indirect-water footprint of this work in Appendix A. The investigation
consumes roughly 2,020 H100-GPU-hours on MareNostrum 5 ACC ---
approximately 1.7 MWh and 225 kg CO\textsubscript{2}eq under a
location-based Scope 2 accounting using Spain's 2025 grid intensity
(Electricity Maps 2025). As a calibration, a single full-scale flagship
model training run (GPT-3 (Patterson et al. 2021)) consumes roughly 750×
the energy of this entire research programme. Null results and
boundary-finding at this scale are a cheap and informative input to the
field's optimisation choices: ruling out a single combination of
(optimiser, schedule, bit-width) interaction at small scale saves
potentially many flagship-training reruns of the same experiment.

\begin{center}\rule{0.5\linewidth}{0.5pt}\end{center}

\subsection{7. Conclusion}\label{conclusion}

We tested the widely-assumed hypothesis that aggressive weight
quantisation requires a different LR schedule than higher-precision
training in sub-100M decoder language models. A 720-cell factorial grid
(Phase 2) at the FP16/INT8/INT6 level falsifies the hypothesis: the
optimal warmdown is 33\% of total iterations regardless of bit-width at
every model size tested.

A 625-cell follow-up programme (Phase 5) probes the robustness of this
null result. We find that it survives changes to the optimiser (Muon →
AdamW), the schedule shape (linear → cosine), the training length (9×
longer), and the size range (down to 5M, up to 350M). The INT6 penalty
follows a log-linear scaling law that predicts five held-out sizes
within their 95\% prediction intervals (5/5).

The INT4 picture, charted across eight model sizes from 3M to 100M, has
two regimes. At 50M and 100M, wd33 is decisively optimal --- every one
of 10 paired-seed comparisons prefers wd33 (paired z ≈ 12--15). Below
50M, across six tested sizes (3M, 5M, 8M, 10M, 15M, 30M), no individual
size shows a statistically significant schedule preference: the per-size
mean penalty oscillates within seed-level noise. The transition between
the two regimes sits between 30M and 50M and is co-located with the
largest single drop in the INT4 quality penalty. This is the first
regime structure we observe in any schedule × bit-width interaction in
our study. A weight-to-grid-distance probe rules out the most obvious
mechanistic explanation for the FP16/INT8/INT6 null result: INT6-QAT
models do not sit closer to the quantisation grid than FP16 models. The
actual mechanism remains open.

The headline practical recommendation is sharper than the Phase 2 null
result alone would license: for sub-100M weight-only QAT at INT8, INT6,
or INT4 at 50M and above, tune the LR schedule once at FP16 and apply
unchanged. For INT4 below 50M, the schedule choice is in the noise ---
wd33 remains a defensible default, and there is no evidence that any
specific alternative consistently outperforms it. The cost of getting
this recommendation wrong is small at INT8 (≤ 0.2 mBPB), modest at INT6
(2--7 mBPB depending on size and training length), and dominated at INT4
by the quality penalty itself (8--12× the INT6 cost) rather than by the
schedule choice.

Two open questions follow naturally and are tractable at small scale.
Does the INT4 noise-dominated regime extend arbitrarily far below 3M, or
does a different schedule eventually emerge at sub-3M scales? And what
is the actual mechanism for the FP16/INT8/INT6 bit-width-agnosticism,
now that grid-snapping has been ruled out? We expect both to be
answerable with an additional few hundred GPU-hours of factorial work in
the same style as Phase 5.

\begin{center}\rule{0.5\linewidth}{0.5pt}\end{center}

\subsection{Acknowledgments}\label{acknowledgments}

This work was conducted at Dwarf A/S, Copenhagen, with portions of the
author's working time supported by Dwarf. Compute was provided by the
EuroHPC Joint Undertaking through AI Factory grant
EHPC-AIF-2026PG01-401, comprising 5,000 GPU-hours on the MareNostrum 5
ACC partition at the Barcelona Supercomputing Center (BSC). We thank
EuroCC Denmark for assistance with the grant application, and the BSC
support team for assistance with the SLURM array job submissions used to
run the 720-cell Phase 2 grid and the 625-cell Phase 5 programme.
Initial development of the training pipeline was conducted as part of
the OpenAI Parameter Golf competition (March--April 2026); the
architectural choices used here originate in that effort.

\begin{center}\rule{0.5\linewidth}{0.5pt}\end{center}

\subsection*{References}\label{refs}
\addcontentsline{toc}{subsection}{References}

\protect\phantomsection\label{refs}

\newpage

\section*{Appendix A --- Computational
Footprint}\label{appendix-a-computational-footprint}
\addcontentsline{toc}{section}{Appendix A --- Computational Footprint}

We report the energy, carbon, and indirect-water footprint of the
experiments in this paper following the systematic-reporting framework
of Henderson et al. (2020) and the methodology of Patterson et al.
(2021). All assumptions are made explicit in Table A.1 so that readers
can substitute their own values and recompute.

\subsection{A.1 Compute accounting}\label{a.1-compute-accounting}

The full investigation --- Phase 2 (the 720-cell factorial grid reported
in §5.1--§5.2) and Phase 5 (the 625-cell follow-up reported in
§5.3--§5.6) --- consumes approximately \textbf{2,020 H100 SXM5
GPU-hours} on the ACC partition of the MareNostrum 5 supercomputer at
the Barcelona Supercomputing Center (BSC) (Banchelli et al. 2025), under
EuroHPC AI Factory grant EHPC-AIF-2026PG01-401. Phase 2 alone consumed
approximately 580 GPU-hours; Phase 5 adds roughly 1,320 GPU-hours of
grid runs plus \textasciitilde10 GPU-hours of pre-flight calibration.

\subsection{A.2 Energy estimate}\label{a.2-energy-estimate}

We estimate total electrical energy consumption \(E\) from GPU-hours
\(H\) as
\[E = H \cdot P_{\text{GPU}} \cdot (1 + \alpha) \cdot \text{PUE}\] where
\(P_{\text{GPU}}\) is the average per-GPU power draw under sustained
training, \(\alpha\) accounts for host overhead (CPU, memory, NVLink,
networking), and PUE is the data centre's Power Usage Effectiveness.
Using the values in Table A.1:
\[E \approx 2{,}020 \cdot 0.600 \cdot (1 + 0.30) \cdot 1.08 = 1{,}702 \text{ kWh} \approx 1.7 \text{ MWh}.\]

\subsection{A.3 Carbon estimate (Scope 2,
location-based)}\label{a.3-carbon-estimate-scope-2-location-based}

Spain's electricity grid had a production-based carbon intensity of
\(133.6\) gCO\textsubscript{2}eq/kWh in 2025 (flow-traced: \(132.1\)
g/kWh) according to Electricity Maps (Electricity Maps 2025). Applying
the location-based figure:
\[\text{CO}_2\text{eq} \approx 1{,}702 \text{ kWh} \cdot 132 \text{ g/kWh} \approx 225 \text{ kg CO}_2\text{eq}.\]

MareNostrum 5's design --- direct hot-water cooling and a stated PUE
target of 1.08 (Banchelli et al. 2025) --- reflects an explicit
efficiency posture, but we could not locate a primary disclosure of
BSC's electricity procurement contracts during the preparation of this
paper. Under a hypothetical market-based Scope 2 accounting that credits
a 100\% renewable supply contract at the lifecycle intensity of
utility-scale solar/wind (\textasciitilde10--50
gCO\textsubscript{2}eq/kWh), the figure would fall to
\textasciitilde17--84 kg CO\textsubscript{2}eq. We report the
location-based figure as the headline; the market-based scenario is
included for completeness only.

\subsection{A.4 Indirect water
estimate}\label{a.4-indirect-water-estimate}

\textbf{We report indirect water only.} Direct on-site cooling water at
the data centre is omitted because BSC has not published a Water Usage
Effectiveness (WUE) figure for MN5 that we could cite. The number below
is therefore a lower bound on total water consumption, capturing
upstream electricity-generation water (primarily thermal-power cooling)
but not the supercomputer's own cooling loop. Using a commonly cited
generic European-grid figure of \textasciitilde1.8 L/kWh:
\[W_{\text{indirect}} \approx 1{,}702 \text{ kWh} \cdot 1.8 \text{ L/kWh} \approx 3{,}060 \text{ L} \approx 3 \text{ m}^3.\]
This is an order-of-magnitude estimate. Spain's grid mix is not strictly
captured by a generic European average --- nuclear remains a substantial
share and is itself water-intensive --- and the absence of a published
BSC WUE means direct on-site cooling water is not accounted for. The
figure should be read as indicative.

\subsection{A.5 Assumptions}\label{a.5-assumptions}

{\def\LTcaptype{none} 
\begin{longtable}[]{@{}
  >{\raggedright\arraybackslash}p{(\linewidth - 4\tabcolsep) * \real{0.3000}}
  >{\raggedleft\arraybackslash}p{(\linewidth - 4\tabcolsep) * \real{0.4000}}
  >{\raggedright\arraybackslash}p{(\linewidth - 4\tabcolsep) * \real{0.3000}}@{}}
\toprule\noalign{}
\begin{minipage}[b]{\linewidth}\raggedright
Symbol
\end{minipage} & \begin{minipage}[b]{\linewidth}\raggedleft
Value
\end{minipage} & \begin{minipage}[b]{\linewidth}\raggedright
Source / rationale
\end{minipage} \\
\midrule\noalign{}
\endhead
\bottomrule\noalign{}
\endlastfoot
\(H\) (GPU-hours) & 2,020 & Tracked by SLURM accounting; Phase 2 + Phase
5 total \\
\(P_{\text{GPU}}\) (W, avg sustained) & 600 & NVIDIA H100 SXM5 has 700 W
TDP; sustained training typically draws \textasciitilde80--90\% of TDP.
A conservative midpoint. \\
\(\alpha\) (host overhead) & 0.30 & Per Patterson et al. (2021);
accounts for CPU + memory + NVLink + networking around the GPU \\
PUE & 1.08 & MareNostrum 5 ACC partition design target; direct hot-water
cooling (Banchelli et al. 2025) \\
Grid carbon intensity (gCO\textsubscript{2}eq/kWh) & 132 & Spain 2025
flow-traced (Electricity Maps 2025) \\
Renewable-PPA scenario (gCO\textsubscript{2}eq/kWh) & 10--50 & Lifecycle
intensity of utility-scale solar/wind \\
Indirect water (L/kWh) & 1.8 & Generic European-grid estimate; no
Spain-specific primary cited \\
\end{longtable}
}

Table A.1 --- Assumptions used in the footprint estimates. Each value is
intended to be replaceable: readers preferring different conventions
(e.g., setting \(P_{\text{GPU}}\) to TDP for an upper-bound estimate)
can recompute the totals trivially.

\subsection{A.6 Context}\label{a.6-context}

For calibration: training a single full-scale flagship model is several
orders of magnitude more energy-intensive than this entire programme.
Patterson et al. (2021) report \textasciitilde1,287 MWh and
\textasciitilde552 t CO\textsubscript{2}eq for the original GPT-3
training. The ratio against this work is \textasciitilde755× in energy
and \textasciitilde2,450× in CO\textsubscript{2}eq (the difference
driven by the relative cleanliness of Spain's grid versus the US data
centre grid used for that benchmark). Small-LM research programmes of
the kind reported here are an inexpensive way to map the optimisation
landscape that flagship model trainings then operate on top of; null
results --- confirmation that a tunable does \emph{not} need
per-precision tuning --- directly reduce the search space of future
flagship-scale runs.

We do not claim this work is ``green''; the impact is non-zero. We do
claim that systematically eliminating hypothesised degrees of freedom at
small scale is a high-information-per-joule activity compared with
re-running the same experiments at larger scale to learn the same thing.

\subsection{A.7 Reporting hygiene}\label{a.7-reporting-hygiene}

Future computational-footprint reporting for follow-up work to this
paper should record the EAR (Energy Aware Runtime) (Banchelli et al.
2025) per-job energy measurements at submission time. EAR is deployed on
the MN5 ACC partition and provides direct measured energy consumption
per SLURM job, replacing the modelled estimates in Table A.1 with
primary data. The estimates here are the best available retroactively
for already-completed runs.

\end{document}